\documentclass{article}  
\usepackage[utf8]{inputenc} 
\usepackage[T1]{fontenc}    
\usepackage{hyperref}       
\usepackage{url}            
\usepackage{booktabs}       
\usepackage{amsfonts}       
\usepackage{nicefrac}       
\usepackage{microtype}      
\usepackage{amsmath}
\usepackage[accepted]{SC2_ICML_i3/icml2019}
\usepackage{color}
\usepackage{amsmath,amssymb,bbm}
\usepackage{enumerate}
\usepackage{graphicx}

\icmltitlerunning{Narration-based Reward Shaping using Grounded Natural Language}

\begin{document}



\twocolumn[
\icmltitle{A Narration-based Reward Shaping Approach \\
            using Grounded Natural Language Commands}



\icmlsetsymbol{equal}{*}

\begin{icmlauthorlist}
\icmlauthor{Nicholas Waytowich}{to,goo}
\icmlauthor{Sean L. Barton}{to}
\icmlauthor{Vernon Lawhern}{to}
\icmlauthor{Garrett Warnell}{to,ed}
\end{icmlauthorlist}

\icmlaffiliation{to}{US Army Research Laboratory, Maryland, USA}
\icmlaffiliation{goo}{Biomedical Engineering Department, Columbia University, NY, USA}
\icmlaffiliation{ed}{Computer Science Department, University of Texas-Austin, Texas, USA}
\icmlcorrespondingauthor{Nicholas Waytowich}{nicholas.r.waytowich.civ@mail.mil}

\icmlkeywords{Machine Learning, ICML}

\vskip 0.3in
]
\printAffiliationsAndNotice{}
\begin{abstract}
While deep reinforcement learning techniques have led to agents that are successfully able to learn to perform a number of tasks that had been previously unlearnable, these techniques are still susceptible to the longstanding problem of {\em reward sparsity}. 
This is especially true for tasks such as training an agent to play StarCraft II, a real-time strategy game where reward is only given at the end of a game which is usually very long.
While this problem can be addressed through reward shaping, such approaches typically require a human expert with specialized knowledge.
Inspired by the vision of enabling reward shaping through the more-accessible paradigm of natural-language narration, we develop a technique that can provide the benefits of reward shaping using natural language commands. Our narration-guided RL agent projects sequences of natural-language commands into the same high-dimensional representation space as corresponding goal states. We show that we can get improved performance with our method compared to traditional reward-shaping approaches. Additionally, we demonstrate the ability of our method to generalize to unseen natural-language commands. 
\end{abstract}

\section{Introduction}
\label{sec:intro}

One of the chief goals in the field of artificial intelligence is to design agents that are capable of solving {\em sequential decision making} problems, i.e, problems in which an intelligent agent is expected not only to make predictions about the world, but also to act within it, and do this continuously over a certain period of time.
Using a class of techniques called {\em reinforcement learning} (RL), artificial agents can, from their own experience, learn solutions to these problems in the form of {\em policies}, or specifications of how they should act.
RL agents attempt to find policies that maximize the amount of {\em reward} they are able to gather, where reward is typically communicated to the agent via a human-specified function that provides a scalar-valued rating of each state the agent may find itself in.
Broadly speaking, RL techniques perform learning by examining the observed reward values and modifying the agent's policy in order to favor repeating those actions that led to positive reward and avoiding those that led to negative reward.

Because the reward values play such a central role during learning, the efficacy of RL techniques is highly dependent on certain qualities of the specified reward function.
One quality of particular importance is referred to as {\em reward sparsity}, which is a measure of how many states in which the designer has specified nonzero (i.e., meaningful) reward values.
Reward functions with fewer nonzero values than others are said to be more sparse, and it is often easier for human designers to specify very sparse reward functions.
For example, if one were to design a reward function for the game of StarCraft II (a complex real-time strategy (RTS) game with fast-paced actions and long time horizons), one could simply set the reward value to be a positive number when the agent has won the game, zero for all other positions or outcomes, and therefore avoid having to think about how to define nonzero rewards for any intermediate game states. 
However, sparse reward functions also negatively impact the efficacy of RL techniques.
Intuitively, this drop in efficacy comes about because the agent receives less meaningful feedback while attempting to perform its task; observed reward values of zero typically lead to no changes to the agent's policy.
In the StarCraft II example, sparse reward functions may lead to the agent spending a considerable amount of time taking random actions in the environment before it happens to win and therefore receive any meaningful feedback.

One class of methods that seeks to address this challenge of reward sparsity is that of {\em reward shaping}.
Reward shaping techniques allow one to modify the agent's reward function in order to encourage faster learning.
Unfortunately, many reward shaping paradigms require that the reward function be modified by humans that have both a certain level of familiarity with how the agent was programmed and the knowledge and access necessary to modify that programming.
In a vision of the future in which autonomous agents serve and team with humans of all sorts, we must enable paradigms of shaping that are accessible to people without this specialized knowledge.

In this paper, we are motivated by the vision of reward shaping through the accessible paradigm of natural-language narration.
By narration, we mean that human designers perform reward shaping not by modifying the source code of the agent, but rather by providing the agent with a sequence of natural-language suggestions for how that agent should go about accomplishing the task.
We propose an extension of a recent reward-shaping technique of this format \cite{Kaplan2017} that uses human-generated game trajectories and natural language command annotations in order to learn an auxiliary reward function that provides extra reward to the agent when a given command has been satisfied.
We extend this technique such that it can be used to enable an RL agent to solve the BuildMarines mini-game in the StarCraft II Learning Environment \cite{Vinyals2017} and, in this experimental context, we are concerned with the following questions:

\textit{1. Is there any benefit in using natural-language-based reward shaping compared to traditional reward shaping?}

\textit{2. To what extent does the proposed reward shaping technique generalize to new, unseen natural language commands?}

Because natural-language-based reward shaping requires extra knowledge (i.e., {\em grounding}, or associating language with state information available to the agent) compared to lower-level shaping approaches, our first question seeks to determine the impact of this requirement on the overall learning process.
We hypothesize that, while the grounding process itself will necessitate more data than low-level reward shaping approaches, natural-language-based shaping itself will not hinder the task-learning process.
Our second question is concerned with the true promise of using natural-language-based shaping techniques, i.e., whether or not what is learned during training will translate to new language suggestions that may be provided by, e.g., other users in the future.
Based on the generalization performance of natural-language representations in other settings, we hypothesize that reward shaping using our technique will also generalize to similar, but different, natural-language shaping suggestions.

The rest of this paper is organized as follows.
We first provide a brief overview of current work in reward shaping in the context of training policies with RL, in addition to current work in RL for Starcraft II.
We then discuss our approach that uses mutual embeddings of natural language and state-action sequences for solving the sub-tasks within the Starcraft II Learning Environment (SC2LE).
Finally, we experimentally investigate the questions we have outlined above, where we confirm our hypotheses--our approach does not hinder task learning and provides a reasonable degree of generalization to new language shaping suggestions.

\section{Related Work}


The ability for RL to learn an optimal policy depends critically on how dense, or frequent, reward signals are provided from the environment. 
For example, it is possible to train an agent policy that can outperform human players for many Atari games when using the game score as a dense reward \cite{Mnih2015a}. However, for games which provide very sparse rewards (i.e.: Montezuma's Revenge and Go) learning is significantly more difficult and thus significant data and computational resources are required to solve for an optimal policy \cite{silver2016}.
One potential approach to alleviate this issue is through \textit{reward shaping} \cite{dorigo1998robot,Mataric1997}, whereby the environment reward function is externally modified, for example by a human observer, to provide more frequent rewards and improve the stability and speed of policy learning. Human observers can provide these reward signal modifications in a multitude of ways, from using demonstrations of the task \cite{Argall2009}, to binary good/bad feedback \cite{Knox2009, Warnell2018} to natural language guidance \cite{Arumugam2017, Matuszek2013, Blukis18, Sung2018, Shah2018, MacGlashan-RSS-15}. In this work we focus on reward shaping using natural language guidance. 

There has been extensive prior work on using natural language based instruction to help autonomous agents learn policies. Application domains range from text-based adventure games \cite{He2016ACL} to learning language-guided autonomous policies for robotic navigation and obstacle avoidance \cite{Mei2016AAAI,Arumugam2017,Matuszek2013,Blukis18,Sung2018,Shah2018, MacGlashan-RSS-15, Artzi2013_ACL, Chen2011_AAAI}. \cite{Fu2018ICLR} proposed a language-conditioned reward learning (LC-RL) framework, which grounded language commands as a reward function represented by a deep neural network to improve learning. 
\cite{Tung_2018_CVPR} collected a narrated visual demonstration (NVD) dataset where human teachers performed activities while describing them in detail. They then mapped the teachers’ descriptions to perceptual reward detectors which they then used to train corresponding behavioural policies in simulation. \cite{Kaplan2017} applied a natural language reward shaping framework to the Atari Game Montezuma's Revenge and showed that it outperformed existing RL approaches that did not use reward shaping. \cite{Blukis18} introduced the Grounded Semantic Mapping Network (GSMN) and applied it to quadrotor navigation in a high-fidelity simulator environment.

There is a growing interest in applying reinforcement learning approaches to multi-agent real-time strategy games such as Starcraft II.
Recent work by \cite{Pang2018} showed that combining hierarchical reinforcement learning (HRL) approach with a hand-crafted reward signal performed better than using a binary reward of 1/0 for win/loss and were able to obtain a $93\%$ win rate against the expert game AI when trained on a computer with 48 cores and 8 Tesla K40 GPUs for two days.
AlphaStar, a Starcraft II agent developed by Google DeepMind, was recently shown to win a series of matches against a human GrandMaster player \cite{alphastarblog}.
They also report that supervised imitation learning on human gameplay data resulted in a policy that could defeat the Elite-level AI approximately $95\%$ of the time. Our work differs from these works in that we focus on natural language guidance to solve for sub-policies which could potentially be used in a HRL framework in a more human-interactive manner. 

\section{Methods and Approach}
\label{sec:meth}
\begin{figure*}[!ht]
	\begin{center}
		\includegraphics[width=0.95\linewidth]{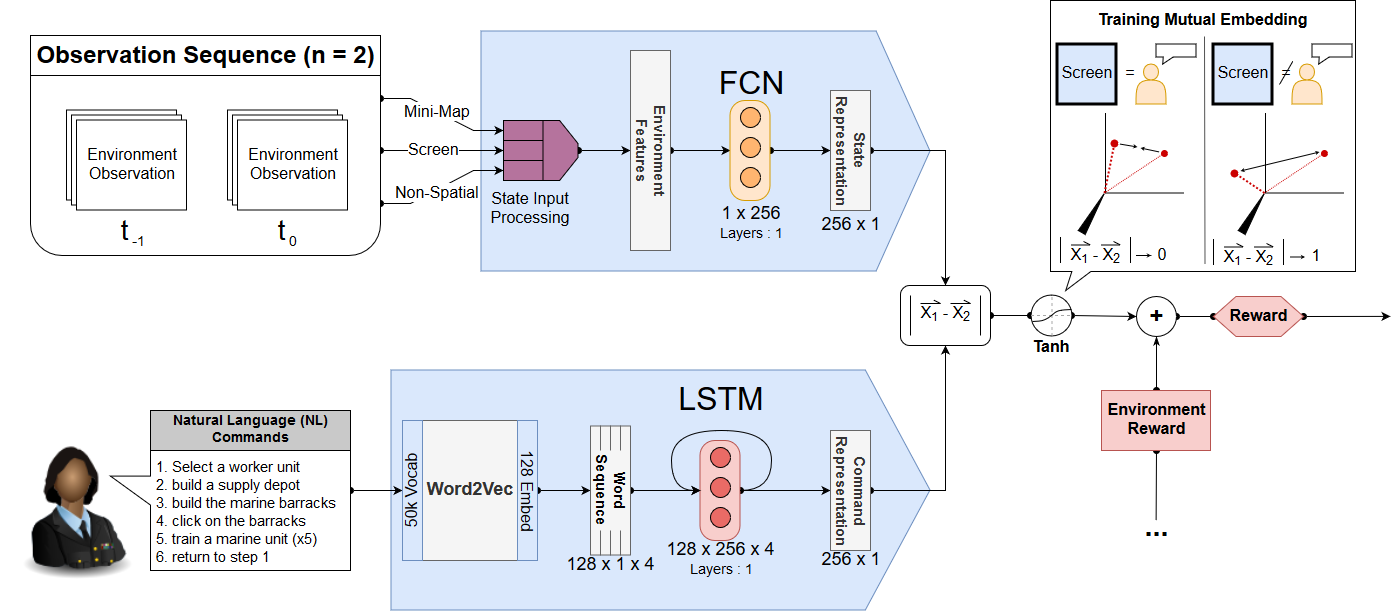}
		\caption{\footnotesize{Diagram of the mutual-embedding model for deriving reward from natural-language commands. The embedding model has two inputs: (Top) the StarCraft game states are fed through the state input processing module and a fully connected layer to transform into a 256 length state embedding. (Bottom) the corresponding natural-language commands are passed through a word2vec encoder and then through an LSTM to end at a 256 length language embedding. The model then learns a common embedding space such that the $\ell_2$-norm of the state and command representations are pushed together if they correspond, or are pushed further apart otherwise (top right).}}
		\label{fig:mutual-embedding}
	\end{center}
\end{figure*}
\subsection{StarCraft II: BuildMarines Mini-game}
In this work we wish to investigate using natural-language commands to train an RL agent to play StarCraft II (SC2). Like most RTS games, StarCraft II is particularly challenging for RL algorithms to solve given the complicated rules, large action spaces, partial-observability of the environment state, long time-horizons and, most of all, due to the sparsity of the game score.  Because of these factors, our first goal was to solve one the simpler, yet still challenging, SC2 mini-games outlined in Vinyals et. al. \cite{Vinyals2017}. There are several mini-games defined, each with varying complexity and difficulty. In this paper, we focus on the most difficult mini-game called BuildMarines \cite{Vinyals2017}. The difficulty of this mini-game arises from its use of a very sparse reward function and as such it remains an open challenge for traditional state-of-the-art RL algorithms. 

As the name suggests, the main goal of this mini-game is to train an agent to build as many marines as possible in a certain time frame. To do this the agent must follow a sequential set of behaviors: build workers, collect resources, build supply depots, build barracks, and finally train marines. The agent starts with a single base and several workers that are automatically set to gather resources, and must learn to construct supply depots (which allow for more controller units to be built), as well as build marine barracks (which allow for marines to be generated) as interim steps before it can achieve its final goal. The agent receives a scalar-valued reward from the environment only when it successfully builds a marine, though it receives additional rewards for each marine built.  In this paper, we reduced the action space from that of the full StarCraft 2 action space to the minimum set of actions to reasonably accomplish the task (see below). 

\subsection{State and Action Spaces for BuildMarines Task}
\begin{figure}[!ht]
	\begin{center}
		\includegraphics[width=1.0\linewidth]{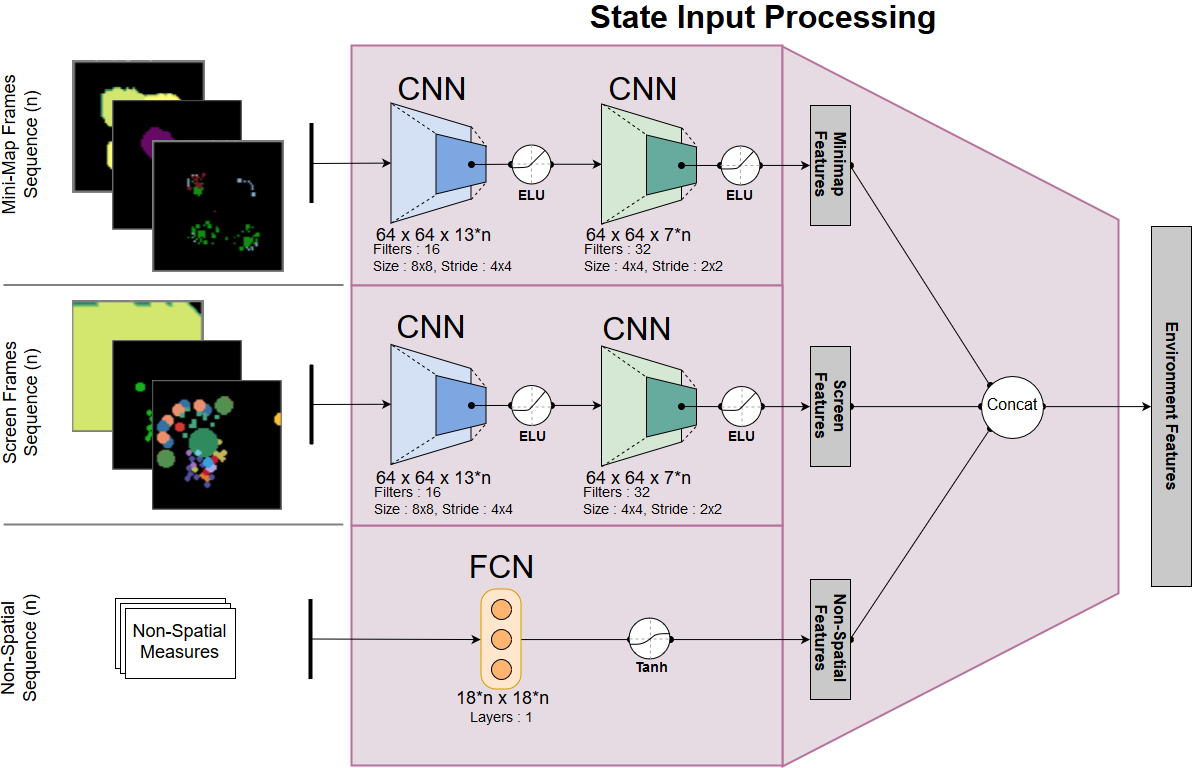}
		\caption{\footnotesize{State input processing. Shown here is the state input processing pipeline for the mutual-embedding model and the A3C agent for the SC2 task. SC2LE provides 3 primary streams of state information: mini-map layers, screen layers, and non-spatial features (such as resources, available actions and build queues). The mini-map and screen features were processed by identical 2-layer CNNs (top two rows) in order to extract visual feature representations of the global and local states of the map, respectively. The non-spatial features were process through a fully-connected layer with a non-linear activation. These three outputs were then concatenated to form the full state-space representation for the agent, as well as for the state-based portion of the mutual-embedding model.}}
		\label{fig:stateInputProcessing}
	\end{center}
\end{figure}

We utilized the StarCraft II Learning Environment (SC2LE) API developed by DeepMind as the primary means for interacting with the StarCraft II (SC2) environment \cite{Vinyals2017}. Using SC2LE, the SC2 state-space consists of 7 mini-map feature layers of size 64x64 and 13 screen feature layer maps of size 64x64 for a total of 20 64x64 2d images (see left panel of Figure \ref{fig:stateInputProcessing}). Additionally, there are 13 non-spatial features that are also part of the state space containing information such as player resources and build queues. These game features were processed using an input processing pipeline, shown in Figure \ref{fig:stateInputProcessing}. The actions in SC2 are compound actions in the form of functions that require arguments and specifications about where that action is intended to take place on the screen. For example, an action such as ``build a  supply depot'' is represented as a function that would require the x-y location on the screen for the supply depot to be built. As such, the action space consists of the action identifier (i.e. which action to run), and an two spatial actions (x and y) that are represented as two vectors of length 64 real-valued entries between 0 and 1. 

\subsection{Narration-Guided RL}
Our goal with this study was to overcome the problem of reward sparsity, wherein the success or failure of a task is determined only by the final state of the environment. Reward sparsity presents a critical problem for learning efficiency because agents must take a enormous number of random actions before stumbling upon the first instance of successful task completion. 


Here, we investigate a narration-guided approach to providing interim rewards to RL agents attempting to solve a complex sequential task with sparse environment rewards. Our narration-guided RL approach consists of two phases: first, we derive reward from natural language by grounding language in terms of interim goal states, then we use that language grounding to shape agent behavior via narrations (or natural-language commands) to guide learning.


\textbf{\textit{Deriving Reward from Natural Language}}: In order for our agent to make use of narrations provided by a human, the language for those narrations needs to be grounded in a context that the agent can understand. In previous work, we developed a \textit{mutual-embedding model} (MEM) to ground natural-language commands and StarCraft II game states in a similar representation space using a multi-input deep neural network and the Word2Vec language embedding model \cite{Waytowich2019}.  In this paper, we build upon this work and use the mutual-embedding model to derive reward from natural language and facilitate our narration-guided RL approach. 
The MEM learns a contextual mapping between the StarCraft 2 game states and a pre-defined set of natural-language instructions (or commands)\footnote{The specific set of natural-language commands that we use in our narration guided approach is shown on the bottom left of Figure \ref{fig:mutual-embedding}.} that indicate the desirable interim goals. The mutual-embedding model (shown in Figure \ref{fig:mutual-embedding} and discussed in detail in the Section 1 of the supplementary material) learns a common representation of the game states and natural-language commands that correspond to those game states. This common representation allows the agent to assign contextual meaning to the current game state that it is experiencing. The model is learned by first projecting the language commands and game states into vector spaces of matching dimensionality, and then minimizing the $\ell_2$-norm between the vectors of corresponding commands and game states, while simultaneously maximizing the $\ell_2$-norm of commands and states that do not correspond. For example, we wish for game-state embeddings that correspond to the command "Build a supply depot" to be closer to that command's vector representation in the mutual embedding space, while being further away from all other command embeddings. The result is a shared embedding space that can represent both the semantic meaning of natural-language commands, as well as the contextual meaning of game states. Ultimately, successfully training the MEM depends on three core processes: the embedding of the natural-language commands, the embedding of the game-states, and the learned correspondence between these two embedding spaces that forms the MEM.


\textbf{\textit{Shaping Agent Behavior with NL}}:
Once the MEM was trained, it was possible to use natural-language commands to provide a form of intuitive reward shaping to a reinforcement learning agent that was attempting to solve the SC2 Build Marines mini-game. natural-language commands corresponding to interim goals (such as ``build a supply depot'') were provided as a form of input to the learning agent. Because the MEM can represent natural-language commands in an embedding space shared by the game states, it is possible for the learning agent to compare its current state in the game with the interim goal state specified by the command. If the agent reaches a game state that satisfies the current instruction (i.e. the normalized euclidean distance from the mutual-embedding model less then some threshold $\tau$), the agent marks that instruction as completed, gives itself an additional positive reward and then moves on to the next instruction. In addition to the reward from the MEM, the command and game state embeddings are also provided to the agent as input. In this way, the mutual embedding functions not only as an internal reward mechanism for the agent, but also provides useful state information that it can learn over.




\begin{figure}[!t]
	\begin{center}
		\includegraphics[width=0.95\linewidth]{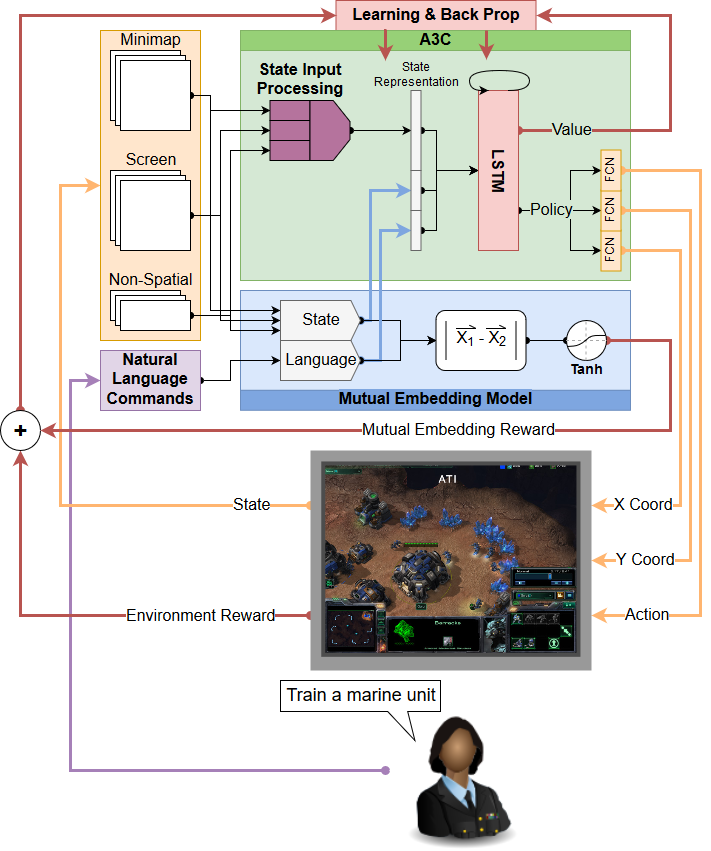}
		\caption{Full narration-guided RL agent model. Shown here is a schematic diagram of the full reinforcement learning agent and its connection to the mutual-embedding model and the environment. As typical of an on-policy agent, the A3C agent here (in green) takes in states and reward information from the task environment, and uses this data to compute actions for the next time step, as well as compute gradients to increment reward maximization. In addition to state information from the environment, our agent also received the state- and command-embedding from the mutual embedding model as inputs to its actor-critic LSTM. Thus, the learned policy and value functions were conditioned not only on the state of the environment, but on the mutual-embedding model's assessment of how close the agent was to the state articulated by the natural-language command. Additionally, the mutual-embedding model augments the environment reward by providing its own reward that reflects how close the agent is to completing the current objective.}
		\label{fig:model}
	\end{center}
\end{figure}

\subsection{Learning Agents}

For all learning agents trained and analyzed here (including our narration guided RL approach) we used an Asynchronous Advantage Actor Critic (A3C) as the core RL algorithm \cite{Mnih2016}. The A3C is an actor-critic method that learns both a value function (the critic) which gives an indication of how much value (or reward) to expect by being in a given state, and a policy (the actor) which maps states to actions to maximize the amount of value (or reward) the agent will experience. 
The A3C is a distributed version of the advantage actor-critic algorithm in which multiple, parallel copies of the actor are created to execute actions and collect experience simultaneously. Having multiple actors collect experience in this way increases the exploration efficiency and thus improves learning (more details can be found in \cite{Mnih2016}).


\textbf{\textit{Narration-Guided RL Agent}}:
For the narration-guided RL agent, we used the pre-trained mutual-embedding model to guide the A3C agent during the learning of the BuildMarines mini-game of SC2. For each parallel worker of A3C agent, an ordered list of instructions is provided for the agent to sequentially complete. As an agent acts in the SC2 environment, it continuously checks to see if the current instruction has been satisfied by passing the current game state it observes as well as the current instruction to complete through the pre-trained mutual-embedding network. If the output of the embedding network is smaller than some threshold $\tau$ ( indicating that the command has been satisfied in terms of game states) the agent marks that instruction as completed, gives itself an additional positive reward (reward-shaping) and moves to the next instruction. In addition to the extra reward that agent gets from the mutual-embedding model, the command embeddings as well as the state embeddings from the mutual embedding model are passed to the A3C agent as additional features to learn from. The intuition is that the agent can learn to use this feature representation to guide itself to completing the next instruction, rather than blindly taking actions until it completes an instruction. The full architecture of the narration-guided RL agent is shown in Figure \ref{fig:model}.

\textbf{\textit{Subtask-Reward Agent (Explicit Reward Shaping)}}:
As a comparison to the narration-guided RL approach, we tested a more explicit form of reward-shaping in which we hand-crafted our own augmented reward-function that we call subtask-reward. We made the subtask-reward module such that it corresponds to the pre-defined set of natural-language commands that we used in our narration-guided approach. The subtask-reward module consists of detectors that identify when the agent has completed one of the subtasks (or commands) and provides additional reward. The detectors for the subtask-reward module are the same detectors used to generate the training dataset for the mutual-embedding model. 
The architecture of the subtask-reward agent is shown in the Supplemental Material in Figure S1.

\textbf{\textit{Baseline agent (no reward shaping)-\cite{Vinyals2017}}}:
As a baseline, we also implemented a standard RL agent with no reward shaping and learns using only the sparse reward provided by the environment. The architecture of baseline agent we use is similar to the Atari-net agent from \cite{Vinyals2017}, which is an A3C agent adapted from Atari to operate on the SC2 state and actions space. We make one slight modification to this agent and add an LSTM layer as it was shown in \cite{Vinyals2017} that adding memory to the model gives improved performance. Both our Narration-guided RL agent and the subtask reward agent use this base architecture at their core (green box in Figure \ref{fig:model} and S1). 
\section{Evaluation}
\label{sec:eval}
In this paper, we sought to evaluate two questions: 1) is there any benefit in using a natural-language-based reward shaping approach compared to traditional reward-shaping, and 2) to what extent does our reward-shaping approach generalize to new, unseen language commands. 
\subsection{Narration-guided RL vs traditional approaches}
To assess the fidelity of natural-language-based reward shaping approach, we first trained the mutual-embedding model to learn an embedding between natural-language commands and game-states and then use this embedding to facilitate reward shaping component of our narration-guided RL agent.
We trained the mutual-embedding model (MEM) using a dataset of 150k samples consisting of pairs of game-states and natural-language commands according to the procedure described in Sections 1.4 and 1.5 in the Supplementary Material.  
Using the trained MEM, we tested the ability of our narration-guided RL agent to solve the BuildMarines mini-game of SC2. Additionally, we trained the sub-task reward agent as well as a standard A3C agent (no reward-shaping) as baselines to compare against our reward-shaping approach. All learning agents were trained for 90 million time-steps using the same hyperparameters. Each agent used 40 parallel workers, an Adam optimizer to minimize the A3C loss (defined in \cite{Mnih2016}), a model rollout length of 200 timesteps and a learning rate of $1e^{-5}$ with exponential decay set to decay every million steps with a base of $0.99$. 

\begin{figure}[!t]
	\begin{center}
		\includegraphics[width=0.95\linewidth]{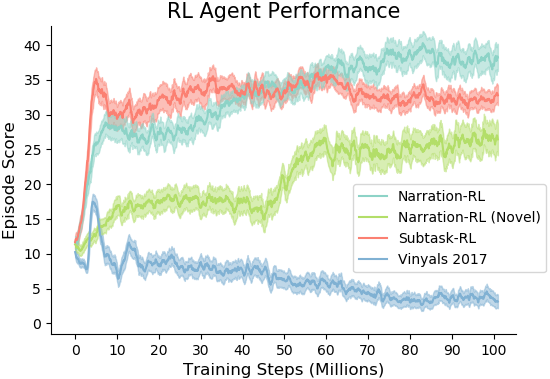}
		\caption{\footnotesize{Average agent performance on the BuildMarines minigame for each of the learning agents tested. Y-axis shows episode score (i.e. \# of marines built per game) and x-axis shows the training step count in millions. Our proposed Narration guided RL agent using the Mutual-Embedding Model (MEM) shown in teal, initially starts with a lower performance compared to the subtask reward shaped agent, shown in red, however the policy gradually improves and eventually outperforms the subtask-rl agent. The standard RL agent (baseline) is shown in blue and a uniform random agent is shown in grey. The green line shows the performance of the narration guided RL agent when given a set of novel, untrained commands. Shaded regions represent standard error.}}
		\label{fig:main_comparison}
	\end{center}
\end{figure}

The training performance of the learning agents is shown in Figure \ref{fig:main_comparison} where the y-axis shows episode score (i.e. \# of marines built per episode) and the x-axis shows the training time in steps. Since the BuildMarines mini-game has a very sparse reward function (i.e the only reward seen is when the objective is completed) we expect the reward-shaping approaches to outperform the standard RL approach. Indeed, we see that both the narration guided and subtask reward shaping agents are able to achieve significantly better performance with over 30 marines created on average compared to the non reward-shaped RL agent (Vinyals 2017). The Vinyals 2017 agent starts off early in the training achieving around 10-12 marines per epsiode and then settles down to around 5 marines per episode at the end of training. Although this is somewhat uncharacteristic of a typical learning curve, the final results do agree with the build-marine results reported by \cite{Vinyals2017} in which they achieved around 5-6 marines on average. 

Examining the performances of the reward-shaped agents, we see that both our Narration-RL and the subtask RL agents experience a rapid improvement in performance in the first 10 million training steps with the subtask-RL agent achieving a higher initial performance of around 30-35 marines compared to the Narration-RL which only achieves around 25-30 on average. After the initial spike in performance, the subtask-RL agent shows a steady plateau in performance for the remainder of the training whereas the Narration-RL agent shows a gradual increase in performance during training and eventually outperforms the subtask-RL agent after around 70 million steps. Overall, we see that our narration-RL agent is able to outperform the traditional subtask-RL agent.

\subsection{Generalizing to Novel Natural Language Commands}
\begin{figure*}[!ht]
    \begin{center}
        \includegraphics[width=0.9\linewidth]{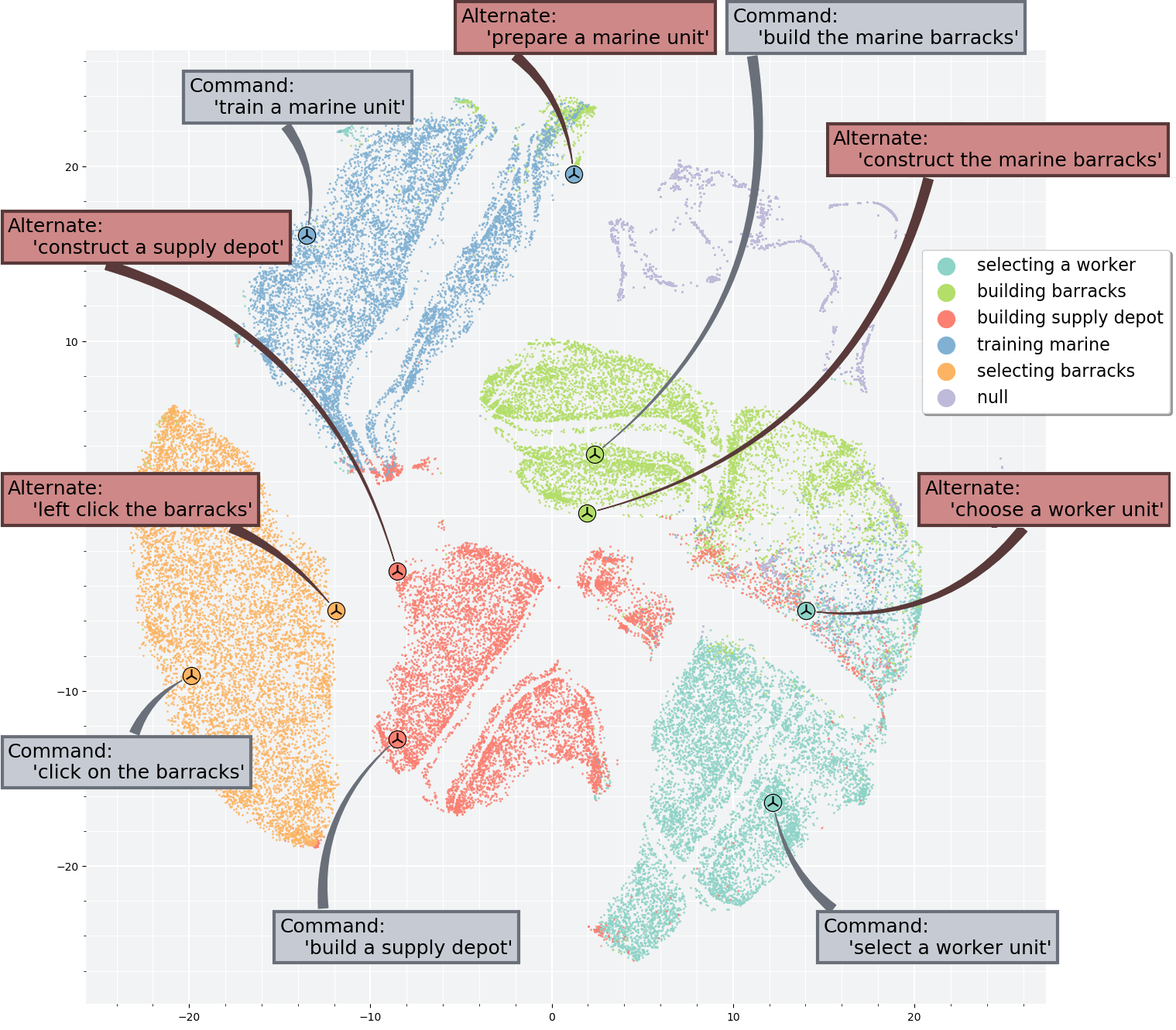}
        \caption{\footnotesize{Learned knowledge representation of the mutual-embedding model. Here we used TSNE to discover if a) the ME model could distinguish between relevant goal states, b) if the MEM learned to project the natural-language commands into the same knowledge representation space as the state representations, and c) if the MEM would generalize to novel NL commands. t-SNE was implemented on the training data used to train the MEM prior to RL implementation. Separation of the state-space examples based on the goal state being demonstrated is clearly visible by the clustering of the data into distinct representation groups. Additionally, the NL commands (blue and labeled "Command") that correspond with each goal state clearly project to the same manifold, demonstrating highly successful learning on the part of the mutual-embedding model. Finally, alternate commands (red and labeled "Alternate") also project to the appropriate regions of the high-dimensional representation space of the MEM.}}
        \label{fig:TSNE}
    \end{center}
\end{figure*}


Using natural language as a means to communicate desired goal states to an agent streamlines the interaction between humans and agent counterparts by leveraging a communication medium that is intuitive and natural for the human user. However, language is highly flexible, and the benefits of the natural ease of using natural language depends in part on how well the communication between a human and an agent maintains this flexibility of language.

For our narration-guided approach, a human user would ideally be able to use an alternate command that is distinct from, yet semantically similar to, the commands used to train the MEM in order to indicate a goal state. Thus the natural language is not just a stand in for another form of selecting a predefined option from a list, but rather provides a flexible way for users to communicate intent to learning agents. In order to test how flexible the MEM model was to variations in natural language, we looked for MEM and RL generalization to a unique set of untrained commands that were semantically similar to the original commands used to train the MEM and RL agents. The specific set of commands used are listed in Table S1 in the supplementary material.

\textbf{\textit{MEM projection generalization to novel commands}}:
We first evaluated generalization by t-distributed stochastic neighbor embedding (t-SNE) \cite{Maaten2008} to visualize how the MEM represented goal states, and whether or not NL commands projected to the appropriate goal state manifolds. We also include the high-dimensional representations of a series of alternate commands that the MEM has never seen in order to test for generalization. If the MEM can successfully generalize the mutual embedding space to semantically similar (but untrained) commands, we should see these distinct but semantically similar commands projecting to the corresponding manifold of goal states in the high-dimensional representation space. These alternate commands were selected by the authors as subjectively intuitive alternatives to the original commands. The goal of this subjective selection process was to produce alternate commands that felt like natural alternatives from a subjective human perspective. These alternate commands were not evaluated for semantic similarity within the context of the utilized Word2Vec embedding network, but such an evaluation could be performed to find even better, more precise command alternatives.

Figure \ref{fig:TSNE} illustrates the results of the t-SNE analysis, as well as the projection of the natural language commands into the MEM's high-dimensional representation space. It is evident that the commands used to train the MEM (and subsequently the RL agent) project to the regions of the representation space where corresponding goal-state representations are grouped. This indicates the MEM successfully maps NL commands into goal-state representations that are grounded in the state information of the BuildMarines SC2 task.

Additionally, we can clearly see that the unseen alternate commands similarly map to the correct corresponding manifold of goal states. This suggests that the MEM is indeed able to generalize the mutual embedding to novel NL commands that have not been observed or learned by leveraging the semantic flexibility of natural language. We can also see how the differences between the original commands and the alternate commands can introduce a degree of ambiguity or uncertainty that could reduce MEM performance and thus RL learning. This is particularly evident for the alternate command "choose a worker unit", which maps to more ambiguous region of the high-dimensional representation space. Implications for this reduced projection precision can be found in the Section 5.

\textbf{\textit{RL performance generalization to novel commands}}:
We also evaluated generalization by testing the performance of our approach when using novel natural language commands during training of the RL agent. We provided our narration-guided RL agent with a set of novel commands that were not previously seen or trained on by the MEM. We used the same alternative set of commands described in the previous section and trained the narration-guided RL agent using the same hyperparameters as before. The generalization performance of our method can be seen in Figure \ref{fig:main_comparison} as the green line labeled 'Narration-RL (novel)'. We see that when using novel language commands during training, our approach does experience reduced performance compared to using the commands to which the MEM was originally trained. However, we still observe that the narration-RL agent is still able to learn the task and achieve an average score of around 25-30, which is still significantly better the RL agent without reward shaping (Vinyals 2017) indicating that we are still gaining some benefit from reward shaping using alternative commands. 

\section{Discussion}
In this work we investigated the benefit of narration-based reward shaping for improving reinforcement learning in the presence of sparse rewards compared to traditional reward shaping approaches. 
Our approach hinges on the ability for natural language commands to be embedded within the same space as high-dimensional task states, such as those used by an RL aglorithm to solve StarCraft II. 
Using this mutual embedding reward shaping approach, we showed improved agent learning compared to a subtask-specific reward shaping model and the baseline agent using no reward shaping on the BuildMarines task of Starcraft II. In addition we showed that the natural language embedding model can still learn in the presence of variations in the natural language input without re-training the mutual embedding model, suggesting the approach can be used across a variety of users who may give narrated commands in slightly different ways. We believe that this aspect is important to improving future Human-AI interactive systems. 

\textbf{\textit{The Generalization of narration-guided learning}}:
We tested the generalization of our approach by utilizing novel language commands during training of the narration-guided RL agent. Although there was a drop in overall performance using the novel language commands, we still observe significant learning on the task indicating the presence of the generalization. This corresponds with the mutual embedding results shown by the t-SNE in Figure \ref{fig:TSNE} where although most of the novel commands project closely to the original commands, some of the novel commands project further away and are thus harder for the MEM to correctly identify. 

Our work used the Word2Vec language embedding model which is a context-free embedding that has been used in previous works. One potential avenue to improve our approach is using more recent techniques such as BERT \cite{Devlin2018_BERT}, a technique based on deep bi-directional recurrent networks that provides context-dependent language embeddings. This will be particularly important as the type of language command being narrated becomes more hierarchical, complex and temporally-dependent.

\textbf{\textit{Limitations and future work}}: 
The performance of any mutual-embedding model using natural language will depend on the diversity and richness of the language descriptions of the tasks being performed. Currently, some works have focused on collecting a large corpus of natural language descriptions for observed tasks and behaviors, such as the LANI dataset \cite{Misra2018-LANI}, which used crowdsourcing to collect language descriptions of drones flying in a high-fidelity simulation environment \cite{Blukis18}. A similar approach could be used here by having human observers provide descriptions of Starcraft II matches.

\bibliographystyle{SC2_ICML_i3/icml2019}
\bibliography{gaw_nrw}

\clearpage

\section{Supplementary Material}

\subsection{Mutual Embedding Model}
\subsubsection{Language Embedding}
To achieve a useful language embedding, we trained a word-level semantic mapping using word2vec with a vocabulary size of 50k and an embedding size of 128. Using this pre-trained word2vec model, word-level semantic embeddings were extracted from each word in a command and then passed through an LSTM model to create a command-level embedding of size 256 (bottom part of Figure 1). The idea behind creating a command level embedding that is derived from word-level semantic embeddings is that it might allow for generalization to new commands composed of words with semantically similar meanings. For example, specific words in a command could be swapped out with semantically similar words (e.g., exchanging ``construct'' for ``build''), and the projection of the new command should be a near-neighbor to the projection of the original command.

\subsubsection{State Embedding}
With the command-level embedding defined, the second stage of the MEM is to project the game states of SC2 into a common embedding space. These game states (composed of mini-map, screen and non-spatial inputs from the SC2LE API) are processed by a state input processing module (shown in Figure 2 of the main paper, which consists of two branches of 2-d convolutional neural networks and fully connected network branch) as a feature extraction step. The SC2 screen and mini-map frames\footnote{A temporal stack of 2 frames (n=2) was used as the game-state input during training of the MEM in order to detect the change that corresponded to meeting the goal.} are each passed through a pipeline of two 2-d convolution layers that extract relevant feature vectors from the images. The non-spatial features are passed through a fully connected layers with a non-linear activation function to create a single non-spatial feature. The three feature outputs are then flattened and concatenated to produce a comprehensive feature array.
Finally, this comprehensive feature layer is projected into a 256-length embedding space using a final fully connected layer, thereby matching the dimensionality of the natural language command embedding.

\subsubsection{Mutual Embedding}
The mutual-embedding model itself (shown in Figure 1 in the main paper) aims to capture a mutual representation of natural language commands and the corresponding game states that serve as interim goals. The model is trained such that game states are pushed closer to their corresponding language commands in the mutual embedding space and are pushed father away from non-corresponding commands. This is done by simultaneously training the embedding networks for both the game states and natural language commands to minimize the $\ell_2$-norm of the difference between the embedding vectors when the game state corresponds to the command, and maximizing the $\ell_2$-norm of the difference between the embedding vectors when the game state and commands do not correspond. The overall loss function used to train is shown below:
\begin{equation}
   \mathcal{L}(\theta) = \frac{1}{N} \sum_n^N \Big(||X_s - X_c|| - y\Big)^2 + \lambda||\theta||^2 
\end{equation}

\noindent where $\theta$ are the neural network embedding parameters, $||\cdot||$ corresponds to the $\ell_2$-norm, $\lambda$ is the $\ell_2$-norm penalty on the network weights ($\lambda = 2.5e^{-3}$ in our case), $X_f$ corresponds to the game state embedding, $X_c$ corresponds to the command embedding and $y \in \{0,1\}$ is the label representing if the game state and command are matching (congruent) or mismatching (incongruent). Our primary objective is to find $\hat \theta = \arg \min_{\theta} \mathcal{L}$, optimized over a set of $n = N$ training samples.

\subsubsection{Dataset Generation for Learning Mutual Embeddings}
To train the mutual-embedding between the natural language commands and the SC2 game states, a labeled dataset is needed that contains pairs of commands and states for supervised learning. In this case, game states corresponding to the different natural language commands need to be collected and there are two main ways this can be done. The first is by using a human (or some other agent) to play the game while following each instruction (or have a human watch the agent play the game) and save the game states when each instruction is reached. However, in order to train a mutual-embedding model consisting of a deep neural network, large numbers of examples are required and thus this option requires a significant amount human effort. The second approach is to use hand-crafted rules to automatically generate or label game states that correspond to each command. Although this option is less burdensome to collect the data, it requires the ability to hand-craft detectors which is not always possible in all tasks or situations. For this paper, we use the second approach to generate a data set since the SC2 state-space is rich enough to construct simple rules that correspond to each instruction. 

For each natural language command, we ran a random agent through the BuildMarines mini-game of the SC2 environment to generate a large set of game states. Then, we use hand-crafted rules that were set up to identify states that satisfy each command to generate corresponding labels. 
An example rule for labeling game states that satisfy the ``build a supply depot'' command, the number of screen pixels corresponding to a supply depot is tracked during the game play of the random agent and whenever that number increases (i.e., the agent has just build another supply depot), then instruction is considered satisfied and the corresponding state is labeled. 

We trained the mutual-embedding model (MEM) using a dataset consisting of pairs of game-states and command embeddings generated by the random agent.  This produced a dataset of 50k game-states consisting of 10k game state examples corresponding to each of the five language commands. We then created matching and mismatching pairs of labeled samples between the states and commands for a total of 100k labeled pairs (50k matched and 50k mismatched pairs). By training on mismatched pairs as well as matched pairs, the model learned not only to associate matched commands and states, but to strongly distinguish commands from other possible game states. The idea here is to learn a mutual embedding manifold where command embeddings are closer to the embeddings of the state they represent while simultaneously being further away from other states. Additionally, we included 50k ``null'' states which were game states that did not correspond to any of the language commands. This null set further distinguished desirable goal states from other states an agent might necessarily pass through. In total, we used a dataset size of 150k samples.

\subsubsection{Training the Mutual-Embedding Model}
To train the MEM, we used the dataset described in the previous section and split the data into training (100k samples), validation (25k samples ) and testing (25k samples) data. 
Training was done using an Adam optimizer with a learning rate of $5e^{-4}$ to minimize the loss function shown in Equation 1. The model was trained using a batch size of 32 over 20 epochs of the data. To prevent overfitting, we chose the point of minimum validation to evaluate our model performance. Our model was able to achieve a training accuracy of  $95.61$ \% a validation accuracy of $82.35$\% and a test accuracy of  $80.40$\%. 

\subsection{Subtask-RL}
\begin{figure}[!ht]
	\begin{center}
		\includegraphics[width=1.0\linewidth]{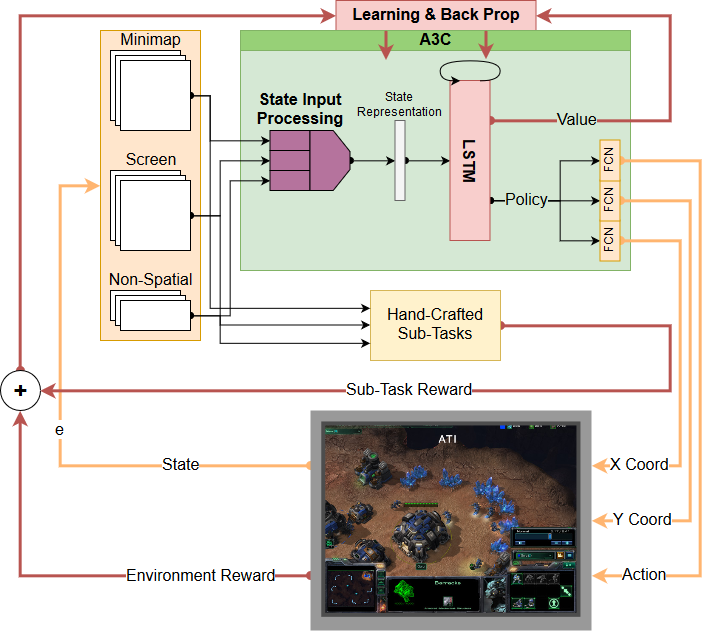}
		\caption{\footnotesize{Subtask based agent model. In contrast to the agent equipped with the mutual embedding model shown in Figure 3 of the main paper, this version of the agent relied entirely on hand-tuned sub-tasks that were defined \textit{a priori} and were based on more traditional conditional representations of intermediate states. These sub-tasks are defined almost entirely on non-spatial features of the task, and as such are not expected to provide as as rich a representation of the state as the mutual-embedding model. Further, they require expert task and programming knowledge to implement, and there is no representation of these sub-tasks in the A3C's state representation.}}
		\label{fig:subtask}
	\end{center}
\end{figure}

We compared our narration-guided RL approach to a traditional reward shaping approach using subtask reward. The subtask-RL model is shown in Figure \ref{fig:subtask}.

\newpage

\subsubsection{Natural Language Commads}

Table \ref{tab:my_label} shows the two sets of Natural language commands used in the Narration guided RL approach. The first is the set of commands used to train the MEM. The second set is used to test the generalization of the MEM to novel, unseen commands. 

\begin{table}[]
    \centering
    \begin{tabular}{c|c} 
         Original Commands & New Commands  \\ \hline
         "select a worker unit" & "choose a worker unit" \\  
        "build a supply depot" & "construct a supply depot" \\ 
        "build the marine barracks" & "construct the marine barracks" \\
        "click on the barracks" & "left click the barracks" \\
        "train a marine unit" & "prepare a marine unit" \\
    \end{tabular}
    \caption{Set of original and novel natural language commands used to train the mutual-embedding model and test it's capacity for generaliztion respectively}
    \label{tab:my_label}
\end{table}

\end{document}